\documentclass[runningheads]{llncs}

\usepackage{eccv}

\usepackage[width=122mm,left=21mm,paperwidth=164mm,height=193mm,top=21mm,paperheight=235mm]{geometry}

\usepackage{eccvabbrv}

\usepackage{graphicx}
\usepackage{booktabs}

\usepackage[accsupp]{axessibility}  

\usepackage{times}
\usepackage{helvet}
\usepackage{courier}
\usepackage[hyphens]{url}
\usepackage[hidelinks]{hyperref}

\usepackage{graphicx}
\usepackage{amsmath,amssymb}
\usepackage{booktabs}
\usepackage{multirow}
\usepackage{microtype}
\usepackage{xcolor}
\usepackage{caption}
\usepackage{subcaption}
\usepackage{algorithm}
\usepackage{algorithmic}
\usepackage{enumitem}
\usepackage{eso-pic}
\DeclareUnicodeCharacter{2194}{\ensuremath{\leftrightarrow}}

\newcommand{\diffR}{\textsc{DiffusionRenderer}}

\usepackage{hyperref}

\begin{document}

\title{HorizonRelight: Relighting Long-horizon Videos Consistently via Diffusion Transformers} 

\titlerunning{HorizonRelight}

\author{Jing Yang\inst{1,2} \and
Mayoore Jaiswal\inst{1} \and
Zian Wang\inst{1} \and
Xiao Steven Zeng\inst{1} \and\\
Yajie Zhao\inst{2} \and
Rochelle Pereira\inst{1} \and
Jianyuan Min\inst{1}\textsuperscript{*}}

\authorrunning{J.~Yang et al.}

\institute{$^{1}$\enspace NVIDIA \quad $^{2}$\enspace University of Southern California \quad \textsuperscript{*}Corresponding author}

\maketitle
\vspace{-12pt}

\begin{center}
    \makebox[\textwidth][c]{\includegraphics[width=1.06\textwidth,trim=0 850 0 0,clip]{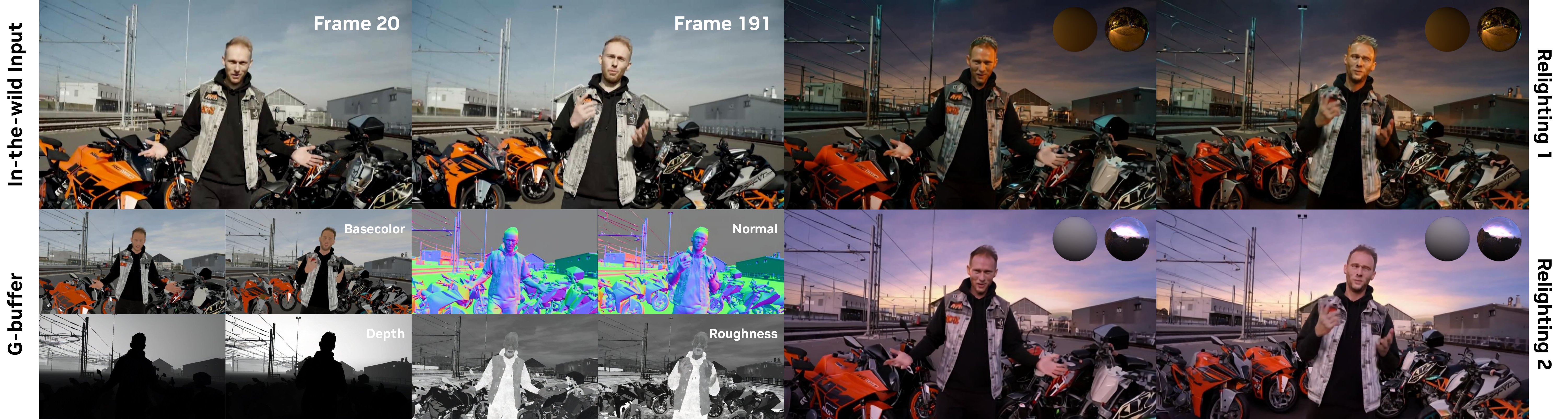}}
    \captionof{figure}{HorizonRelight delivers temporally consistent long-horizon relighting under a target illumination from a single input video by coupling inverse decomposition with forward re-synthesis, with inverse-estimated G-buffer conditioning preserving content consistency over long horizons.}
    \label{fig:teaser}
    \vspace{-10pt}
\end{center}

\begin{abstract}
Diffusion-based video relighting enables controllable relighting from a single input video, but modern video diffusion backbones are trained on short clips and applied to long-horizon videos through chunked sliding-window inference, often causing temporal discontinuities at chunk boundaries. We address this by reframing long-horizon relighting as \emph{temporally conditioned latent domain translation}. Our framework enforces cross-chunk continuity by propagating target-domain latents across boundaries and makes this behavior learnable using \emph{masked target-domain self-conditioning}, training the model to continue from temporally masked propagated context. We further introduce \emph{warm-start prompting} with a relit prompt anchor from a controllable generative model, which establishes the initial target-domain state and creates a general interface for prompt-based relighting. Experiments on in-the-wild long-horizon videos show markedly improved temporal consistency, with chunk-boundary artifacts largely reduced and unwanted appearance changes across chunks greatly suppressed. Visit our webpage at \href{https://research.nvidia.com/labs/sil/projects/horizonrelight/}{\textcolor{eccvblue}{{\urlstyle{same}\nolinkurl{https://research.nvidia.com/labs/sil/projects/horizonrelight/}}}}.
\vspace{-9pt}
\end{abstract}


\section{Introduction}
\label{sec:intro}

Relighting is a core capability for modern visual production, enabling post-capture lighting edits for content creation, AR/VR, and visual effects.
Recent advances combine inverse decomposition with forward re-synthesis by leveraging video diffusion priors, making controllable relighting possible from a single input video \cite{DiffusionRenderer,he2025unirelight}.
In particular, \diffR{} explicitly models both inverse decomposition and forward re-synthesis within a unified diffusion-based pipeline \cite{DiffusionRenderer}, while UniRelight performs end-to-end relighting in a single model \cite{he2025unirelight} and already delivers strong results on short clips on the order of a few seconds (e.g., $\sim$2\,s).
In film practice, takes/shots often span only a few to a few tens of seconds (average shot length roughly 5--25\,s across eras)~\cite{cutting2011quicker}, which already pushes beyond the fixed clip lengths that current diffusion backbones can process reliably at high resolution.

Modern video diffusion backbones rely on spatiotemporal self-attention, whose memory grows rapidly with sequence length and resolution~\cite{ma2025latte,zhang2025faster}.
As a result, training is typically performed on fixed-length clips (``chunks'') that fit GPU memory, and long-horizon video inference is implemented as a sliding-window procedure over time.
This is not a modeling preference but a compute constraint.
Even systems intended for long-horizon prediction often run as bounded-window continuations to efficiently use GPU resources~\cite{po2025long}.
Distributed strategies such as context parallelism further reinforce this regime by scaling effective context across GPUs while keeping per-device windows bounded \cite{jiang2025dcp}.
In the relighting setting, public Cosmos-based diffusion-renderer models are explicitly framed around 57-frame clips, reflecting the practical fixed-window deployment mode \cite{DiffusionRenderer,alhaija2025cosmos}.

\begin{figure}[t]
    \centering
    \includegraphics[width=\linewidth,trim=0 850 0 0,clip]{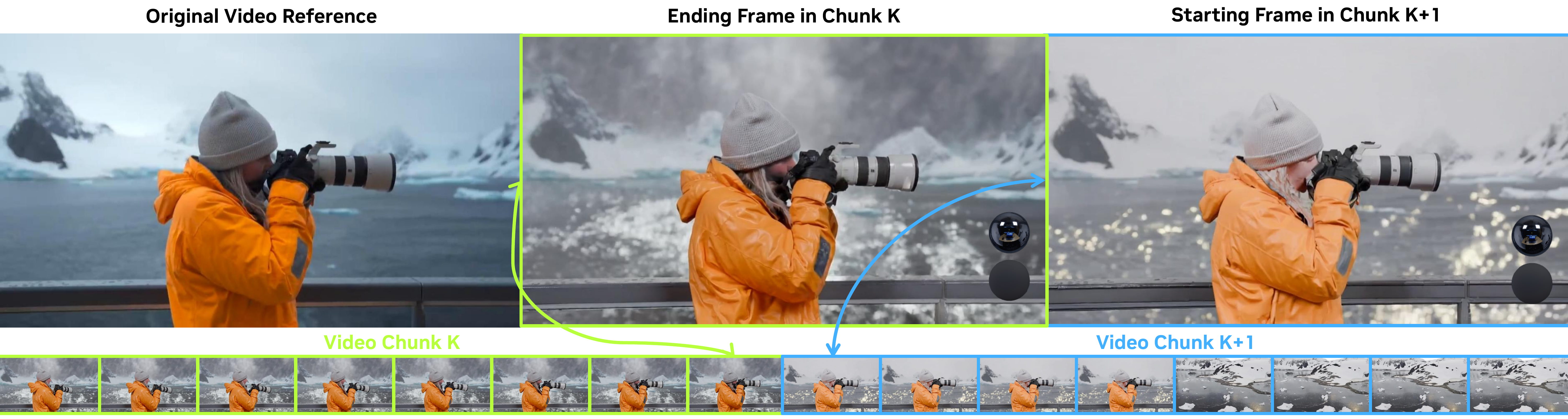}
    \caption{\textbf{Temporal inconsistency under chunked long-horizon relighting.}
    Left to right: input frame, then two relit results produced under the same target illumination in two independent chunked runs via~\cite{DiffusionRenderer}.
    The mirror ball and gray ball indicate the intended illumination, yet the visible differences across runs are enough to break long-horizon consistency.}
    \vspace{-15pt}
    \label{fig:temporal_inconsistency}
\end{figure}

In the compute-constrained chunked regime, relighting pipelines process each chunk largely independently, sometimes with simple overlap or blending~\cite{DiffusionRenderer,he2025unirelight}.
While each window can look sharp in isolation, independence forces the model to re-infer latent causes such as intrinsics and appearance cues at every chunk, creating a train--test mismatch between short-clip training and long sliding-window deployment.
As a result, visible differences can appear whenever the window shifts, leading to temporal discontinuities near chunk boundaries and making relighting unreliable for long-horizon videos.
\Cref{fig:temporal_inconsistency} visualizes this temporal inconsistency under chunked execution in \diffR{}~\cite{DiffusionRenderer}.

In this work, we address chunk-boundary inconsistency by reframing long-horizon relighting as \emph{temporally conditioned latent domain translation}.
Rather than treating each sliding-window chunk as a fresh short-clip generation problem, we model long-horizon relighting as a chained continuation process in the relit target domain.
At inference time, each chunk is conditioned on the previous chunk’s ending target-domain latents, so the relit state is continued rather than repeatedly re-inferred, reducing boundary discontinuities and long-horizon drift.
To make this continuation behavior learnable, we train with \emph{masked target-domain self-conditioning}, which temporally masks propagated target-domain context during training so the model learns to continue relighting from partial evidence.
This learned continuation also enables a principled initialization for the first chunk, where no propagated target-domain state exists; we therefore introduce \emph{warm-start prompting}.
We initialize the forward stage with an externally generated relit prompt anchor, which can be either a single image or a short video from a controllable generative model (e.g., ChatGPT~\cite{achiam2023gpt}, Gemini~\cite{team2023gemini}).
This establishes the initial target-domain state for the subsequent chunk chain and exposes a direct interface for prompt-based relighting edits.
We evaluate on in-the-wild long-horizon videos and show that the proposed training and inference yield markedly improved temporal consistency, with chunk-boundary artifacts largely removed and unwanted appearance changes across chunks greatly reduced.
\section{Related Work}
\label{sec:related}

\subsection{Video Relighting}
\label{sec:rw_relighting}

Relighting aims to modify scene illumination while preserving scene content and material appearance.
A classical line of work approaches this problem through explicit inverse rendering and scene reconstruction, where geometry, reflectance, and lighting are estimated and then combined with physically-based rendering under novel illumination.
This paradigm includes calibrated capture systems such as light stages~\cite{debevec2000acquiring} as well as more recent neural inverse-rendering and neural rendering pipelines built on explicit scene representations.
While these methods provide strong physical interpretability and control, they often rely on multi-view observations, static-scene assumptions, or per-scene optimization, which limits their applicability to in-the-wild dynamic videos.

Recent learning-based methods instead seek to learn relighting across scenes from data.
A representative direction uses intrinsic buffers or latent scene attributes as intermediate structure priors for controllable image synthesis.
For example, Deep Shading~\cite{nalbach2017deep} demonstrated that learned deferred shading can map G-buffers to realistic appearance, while RGB$\leftrightarrow$X~\cite{zeng2024rgb} unified intrinsic decomposition and image synthesis in a diffusion-based framework.
Building on this line, DiffusionRenderer~\cite{DiffusionRenderer} extends diffusion-based inverse and forward rendering to video, using an explicit two-stage pipeline that first predicts editable scene attributes and then synthesizes relit outputs conditioned on those attributes and the target lighting.
This explicit decomposition improves controllability, but it also makes the forward stage sensitive to errors in the intermediate buffers.

Another recent direction uses diffusion models to perform relighting more directly.
Prior works such as LightIt~\cite{kocsis2024lightit}, DiLightNet~\cite{zeng2024dilightnet}, Neural Gaffer~\cite{jin2024neural}, GenLit~\cite{bharadwaj2025genlit}, IC Light~\cite{zhang2025scaling}, and RelitLRM~\cite{zhang2024relitlrm} show that modern generative models can synthesize challenging illumination effects such as cast shadows, specular highlights, and reflections.
However, many of these methods are designed for narrower domains, such as portraits, isolated objects, or sparse-view object reconstruction.
UniRelight~\cite{he2025unirelight} moves toward a more general video setting by jointly modeling relit appearance and albedo in a single diffusion model, avoiding explicit G-buffer bottlenecks and reducing error accumulation between inverse and forward stages.

\subsection{Long-horizon Video Diffusion and Chunked Generation}
\label{sec:rw_longvideo}

Modern video diffusion models achieve strong visual quality by operating in a compressed latent space, but their spatiotemporal computation remains expensive.
As a result, practical video diffusion systems are typically trained and deployed on bounded clips rather than arbitrarily long sequences.
This operating regime underlies latent video diffusion backbones such as Stable Video Diffusion~\cite{blattmann2023stable} and newer DiT-based video foundation models such as Cosmos~\cite{agarwal2025cosmos}.
Related video-generation work has explored stronger multi-modal and motion-aware representations, for example by jointly modeling multiple targets in a single diffusion transformer~\cite{lu2025matrix3d,chefer2025videojam}.
However, these methods do not directly address relighting consistency when a long video must be generated sequentially across chunk boundaries.

Existing video relighting systems inherit the same bounded-context regime.
DiffusionRenderer~\cite{DiffusionRenderer} formulates relighting through inverse and forward video diffusion stages, while UniRelight~\cite{he2025unirelight} also builds on a clip-based latent video diffusion backbone.
These methods produce strong short-clip results, but they do not explicitly address the deployment setting in which long videos must be processed as a sequence of chunks.
In that regime, weak coupling across chunks can lead to repeated re-inference of target-domain appearance and visible discontinuities at chunk boundaries.

\subsection{Prompting and Model Adaptivity}
\label{sec:rw_prompting}

Beyond architecture and clip length, another relevant perspective is how pretrained models are conditioned at inference time. In long-horizon video generation under chunked execution, each chunk is produced under incomplete temporal context, making the quality of the available prefix especially important. This connects naturally to a broader literature on prompting and prefix-conditioned behavior, which studies how pretrained models can be steered by conditioning on an explicit prefix rather than by modifying the model itself.

Decoder-only autoregressive language models (e.g., GPT-style models) are widely studied for prompt-based adaptation, where a prefix can steer the model toward a downstream behavior without changing the architecture or updating parameters at inference time~\cite{brown2020language}. This view is further reinforced by later work on prompt-based and prefix-based adaptation, which shows that conditioning a frozen model with learned prompts or continuous prefixes can serve as an effective alternative to full fine-tuning~\cite{li2021prefix,lester2021power}. By contrast, encoder-only models (e.g., BERT-style models) are pretrained for bidirectional representation learning through objectives such as masked-token prediction, and are typically adapted by attaching task-specific output heads rather than by open-ended continuation from a prompt~\cite{devlin2019bert}. This distinction in training objective and adaptation pattern is commonly used to explain why decoder-only models exhibit strong in-context controllability, while encoder-only models are more naturally used for feature extraction and discriminative transfer.
\section{Preliminaries}
\label{sec:prelim}

\subsection{Video Diffusion Models}
\label{sec:video_diffusion}

We build on a latent video diffusion model (VDM) based on the Diffusion Transformer (DiT) architecture~\cite{Peebles_2023_ICCV}, following the Cosmos formulation adopted by recent video relighting work~\cite{alhaija2025cosmos,he2025unirelight}. Given an RGB video clip $I \in \mathbb{R}^{F \times H \times W \times 3}$, a pretrained VAE encoder $E$ maps it to a latent tensor $z = E(I) \in \mathbb{R}^{F' \times h \times w \times C}$, and a pretrained decoder $D$ maps the latent back to video space. As in UniRelight~\cite{he2025unirelight}, we use a Cosmos tokenizer, namely \textit{Cosmos-1.0-Tokenizer-CV8x8x8}, that compresses the video along both temporal and spatial dimensions, so typically $F' = F/8$, $h = H/8$, $w = W/8$ and, $C=16$. In our setting, the denoiser is a DiT initialized from a Cosmos video diffusion backbone and fine-tuned for relighting~\cite{alhaija2025cosmos,he2025unirelight}.

The diffusion process is performed in latent space under the EDM parameterization~\cite{karras2022elucidating}. Given a clean latent $z_0$, a noisy latent $z_t$ is constructed by adding Gaussian noise according to the EDM noise schedule. A conditional denoiser $f_\theta$ predicts the injected noise from $z_t$ under task-specific conditioning inputs $\mathrm{cond}$. We optimize the standard conditional denoising objective
\begin{equation}
\mathcal{L}_{\mathrm{VDM}}
=
\mathbb{E}_{z_0,\epsilon,t}
\left[
\left\|
\epsilon - f_\theta(z_t \mid \mathrm{cond})
\right\|_2^2
\right],
\label{eq:prelim_vdm_obj}
\end{equation}
where $\epsilon \sim \mathcal{N}(0, I)$. At inference time, the model iteratively denoises from Gaussian noise in latent space and decodes the final latent with $D$.

\subsection{Chunked Execution for Long Videos}
\label{sec:chunked_regime}

Due to the high cost of modeling long spatiotemporal token sequences, DiT-based video diffusion models~\cite{Peebles_2023_ICCV,alhaija2025cosmos,he2025unirelight} are typically trained and deployed with a fixed temporal window.
Given a long video $I^{1:T}$, we partition it into $K$ clips $\{I_k\}_{k=1}^{K}$, where each $I_k \in \mathbb{R}^{F \times H \times W \times 3}$ denotes a bounded-length chunk processed under this fixed clip budget.

This fixed-window regime is the practical setting for long-horizon video relighting, but it provides only limited temporal coupling across chunk boundaries.
As a result, when a long video is handled as a sequence of bounded-window generations, the relit target-domain state may be re-established at each window shift, leading to visible mismatch near chunk boundaries.
In the following section, we specify how the conditioning term $\mathrm{cond}$ in Eq.~\eqref{eq:prelim_vdm_obj} is augmented to support cross-chunk continuation.
\section{Method}
\label{sec:method}

\subsection{Problem Formulation}
\label{sec:formulation}

Our objective is to achieve temporally consistent relighting for long-horizon videos. Let the input video be $I^{1:T} \in \mathbb{R}^{T \times H \times W \times 3}$ and let $\ell^{1:T}$ denote the aligned per-frame lighting condition, represented as equirectangular environment maps. We partition the sequence into $K$ non-overlapping chunks $\{(I_k,\ell_k)\}_{k=1}^K$, where each chunk has length $F$. The task is to generate a relit long video $\hat{I}^{1:T}$ by applying a chunked relighting model $\mathcal{G}_\theta$ sequentially, i.e., $\hat{I}_k = \mathcal{G}_\theta(I_k,\ell_k)$ for $k=1,\dots,K$.

A key requirement is that neighboring predictions remain consistent across chunk boundaries. We therefore view long-horizon relighting as a \emph{consistent long-horizon latent domain translation} problem, where the last $b$ frames of $\hat{I}_k$ should agree with the first $b$ frames of $\hat{I}_{k+1}$ in both content and appearance. Formally, letting $\mathrm{Tail}(\cdot)$ and $\mathrm{Head}(\cdot)$ extract the last and first $b$ frames, respectively, we seek to minimize the seam discrepancy $\Delta_{\mathrm{seam}}(\mathrm{Tail}(\hat{I}_k), \mathrm{Head}(\hat{I}_{k+1}))$ for all adjacent chunk pairs.

\subsection{Propagated Context for Consistent long-horizon Latent Domain Translation}
\label{sec:longtake_generic}

\begin{figure}[t]
  \centering
  \includegraphics[width=\linewidth,trim=0 800 0 0,clip]{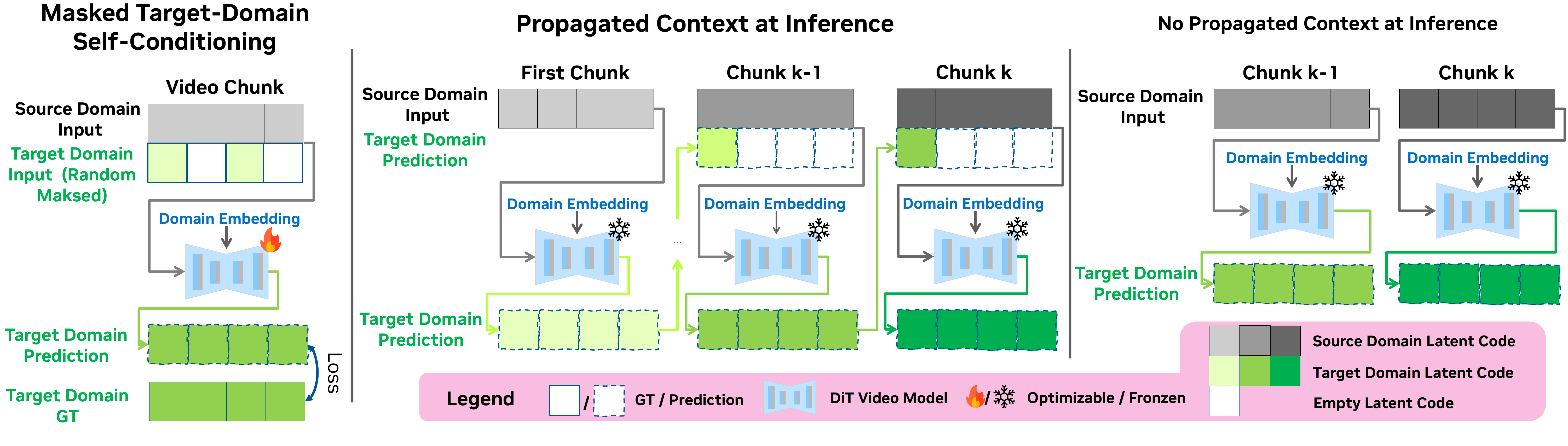}
  \caption{\textbf{Masked target-domain self-conditioning and propagated context.}
  During training, a subset of target-domain latent frames is retained as observed anchors while the remaining frames are masked out and predicted. At inference, the observed anchors are provided by the ending latent frames from the previous chunk prediction, enabling chained generation across chunks.}
    \vspace{-10pt}
  \label{fig:prop_context}
\end{figure}

For a sequence of chunks, we denote the source-domain latent chunks by $\{z^I_k\}_{k=1}^K$, where $z^I_k = E(I_k) \in \mathbb{R}^{F' \times h \times w \times C}$, and the corresponding target-domain latents by $\{z^Y_k\}_{k=1}^K$. We achieve cross-chunk consistency by training the model to predict target-domain latents from masked target-domain observations, and then reusing the same conditioning structure across chunks at inference time.

\paragraph{Masked Target-domain Self-conditioning.}
During training, we condition the model on the source-domain latent and a masked target-domain latent, and train it to reconstruct the full target-domain latent, as illustrated in Fig.~\ref{fig:prop_context} (left). Concretely, for paired latent chunks $(z^I_k, z^Y_k)$, we sample an arbitrary temporal mask $m \in \{0,1\}^{F'}$, where $m_t=1$ indicates an observed frame and $m_t=0$ indicates a masked-out frame, and broadcast it to the latent shape, yielding $M \in \{0,1\}^{F' \times h \times w \times C}$. We then form the masked target-domain latent as $\bar{z}^{Y}_k = M \odot z^Y_k$, so that only observed target-domain entries are retained. The conditioning input is defined as $\mathrm{cond} = \langle z^I_k,\; \bar{z}^{Y}_k \rangle$, yielding
\begin{equation}
\mathcal{L}_{\mathrm{VDM}}
=
\mathbb{E}_{z^Y_k,\epsilon,t}
\left[
\left\|
\epsilon - f_{\theta}(z_t \mid \langle z^I_k,\; \bar{z}^{Y}_k \rangle)
\right\|_2^2
\right],
\label{eq:masked_selfcond_vdm_obj}
\end{equation}
where $z_t$ is the noised version of the target-domain latent $z^Y_k$.

\paragraph{Cross-chunk Context Propagation and Chained Inference.}
At inference, the same conditioning structure is applied sequentially across chunks, as illustrated in Fig.~\ref{fig:prop_context} (middle). For chunk $k$, we use the previous prediction as the target-domain anchor for the current chunk, with propagated context $s_{k-1} = \mathrm{Tail}(\hat{z}^{Y}_{k-1})$, where $\mathrm{Tail}(\cdot)$ extracts the last $b$ frames along the latent temporal axis. For the first chunk, the initial context $s_0$ can be provided by an external starter anchor; if instead $s_0=\varnothing$, the generation is cold-started and may degrade in visual fidelity. These propagated frames serve as target-domain anchors for the next chunk, and the model predicts the remaining target-domain frames conditioned on this partial context. In practice, adjacent chunks can additionally be processed with overlapping frames to further smooth transitions near chunk boundaries. After generating chunk $k$, we update the propagated context as $s_k=\mathrm{Tail}(\hat{z}^{Y}_k)$ and use it to condition the next chunk, enabling chained inference across the full sequence.

\subsection{Relighting Long-horizon Videos with Warm-start Prompting}
\label{sec:relight_longtake}

\begin{figure}[t]
  \centering
  \includegraphics[width=\linewidth,trim=0 500 0 0,clip]{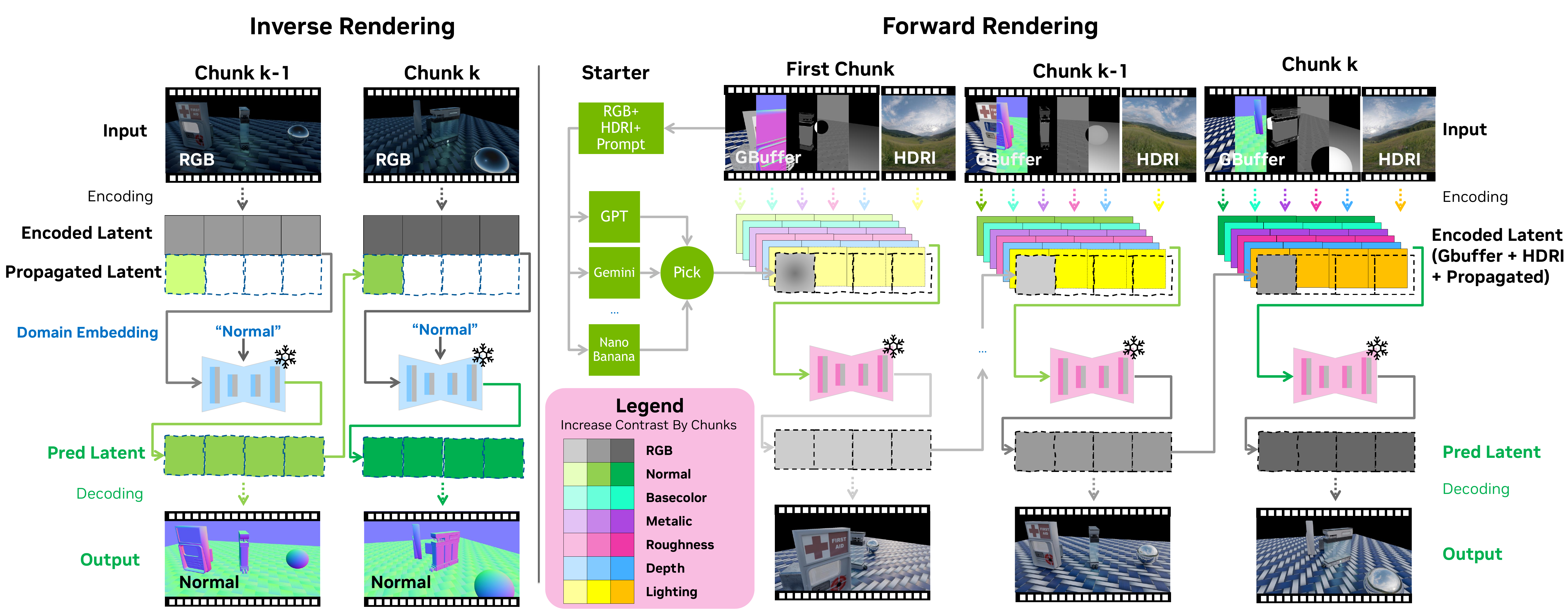}
  \caption{\textbf{Applying propagated context for consistent long-horizon relighting.}
  The propagated context is taken as the ending latent frames from the previous chunk prediction and used to condition the next chunk. We apply the same mechanism to both stages: inverse rendering in the intrinsic latent domains and forward rendering in the relit RGB latent domain.}
    \vspace{-10pt}
  \label{fig:prop_context_dr}
\end{figure}

Following the DiffusionRenderer architecture~\cite{DiffusionRenderer}, we define long-horizon relighting with an inverse stage $\mathcal{F}^{\mathrm{inv}}_{\theta}$ and a forward stage $\mathcal{F}^{\mathrm{fwd}}_{\phi}$:
\begin{equation}
\hat{X}_k = \mathcal{F}^{\mathrm{inv}}_{\theta}(I_k \mid \mathrm{cond}^{\mathrm{inv}}),\qquad
\hat{I}^{\ell}_k = \mathcal{F}^{\mathrm{fwd}}_{\phi}(\hat{X}_k,\; \ell_k \mid \mathrm{cond}^{\mathrm{fwd}}),
\label{eq:dr_stages}
\end{equation}
where $\hat{X}_k$ is the predicted intrinsic sequence for chunk $k$, and $\hat{I}^{\ell}_k$ is the relit RGB output. We use $\mathcal{D}$ for the intrinsic domains, including \texttt{basecolor}, \texttt{normal}, \texttt{metallic}, \texttt{roughness}, \texttt{depth}, and \texttt{specular}. In latent space, $z^I_k$ denotes the input RGB latent, $z^{X_d}_k$ the intrinsic latent for domain $d \in \mathcal{D}$, and $z^{I^\ell}_k$ the relit RGB latent. Here, $\mathrm{cond}^{\mathrm{inv}}$ and $\mathrm{cond}^{\mathrm{fwd}}$ denote the stage-specific conditioning inputs used in each stage. We visualize the long-horizon relighting architecture in Fig.~\ref{fig:prop_context_dr}.

\paragraph{Training.}
During training, both stages operate on fixed-length video clips with masked target-domain self-conditioning. For each intrinsic domain $d \in \mathcal{D}$, we use the corresponding domain embedding $e_d$ to specify the target intrinsic domain, and form a masked intrinsic latent $\bar{z}^{X_d}_k = M \odot z^{X_d}_k$, yielding $\mathrm{cond}^{\mathrm{inv}} = \langle z^I_k,\; e_d,\; \bar{z}^{X_d}_k \rangle$ and
\begin{equation}
\mathcal{L}_{\mathrm{VDM}}^{\mathrm{inv}}
=
\mathbb{E}_{z^{X_d}_k,\epsilon,t}
\left[
\left\|
\epsilon - f^{\mathrm{inv}}_{\theta}(z_t \mid \langle z^I_k,\; e_d,\; \bar{z}^{X_d}_k \rangle)
\right\|_2^2
\right].
\label{eq:inv_vdm_obj}
\end{equation}
For the forward stage, let $\ell_k$ denote the target lighting condition, represented in the same form as prior video relighting frameworks~\cite{DiffusionRenderer,he2025unirelight}. We then form a masked relit latent $\bar{z}^{I^\ell}_k = M \odot z^{I^\ell}_k$, yielding $\mathrm{cond}^{\mathrm{fwd}} = \langle \{z^{X_d}_k\}_{d\in\mathcal{D}},\; \ell_k,\; \bar{z}^{I^\ell}_k \rangle$ and
\begin{equation}
\mathcal{L}_{\mathrm{VDM}}^{\mathrm{fwd}}
=
\mathbb{E}_{z^{I^\ell}_k,\epsilon,t}
\left[
\left\|
\epsilon - f^{\mathrm{fwd}}_{\phi}(z_t \mid \langle \{z^{X_d}_k\}_{d\in\mathcal{D}},\; \ell_k,\; \bar{z}^{I^\ell}_k \rangle)
\right\|_2^2
\right].
\label{eq:fwd_vdm_obj}
\end{equation}
Here, $M$ is sampled by an arbitrary temporal mask and $z_t$ denotes the noised target latent in the corresponding domain.

\paragraph{Warm-start Prompting.}

Warm-start prompting follows the same intuition as prompt-based adaptation in GPT-style models: a prefix defines the task and steers generation without fine-tuning the model itself.
In the relighting setting, the prefix is a relit starter anchor, provided as a single image or a well-aligned short video (e.g., a relit clip from~\cite{DiffusionRenderer}).
This starter anchor fits naturally into our continuation design: it initializes the chunk chain by defining the desired target-domain appearance that the model should start from and continue under.
In the inverse stage, we estimate intrinsic factors and G-buffers for forward-stage content conditioning, with the first chunk also cold-started.
In our forward stage, the relit latent anchor initializes the target-domain appearance, the predicted intrinsic latents $\{z^{X_d}_k\}_{d\in\mathcal{D}}$ preserve scene content, and the lighting condition $\ell_k$ specifies the target illumination.
The starter anchor serves as a visual prompt that initializes the chain, after which generation proceeds under the same continuation process, and it can be produced by any controllable external model.


\begin{figure}[t!]
    \centering
    \vspace{-10pt}
    \includegraphics[width=\linewidth,trim=0 425 400 0,clip]{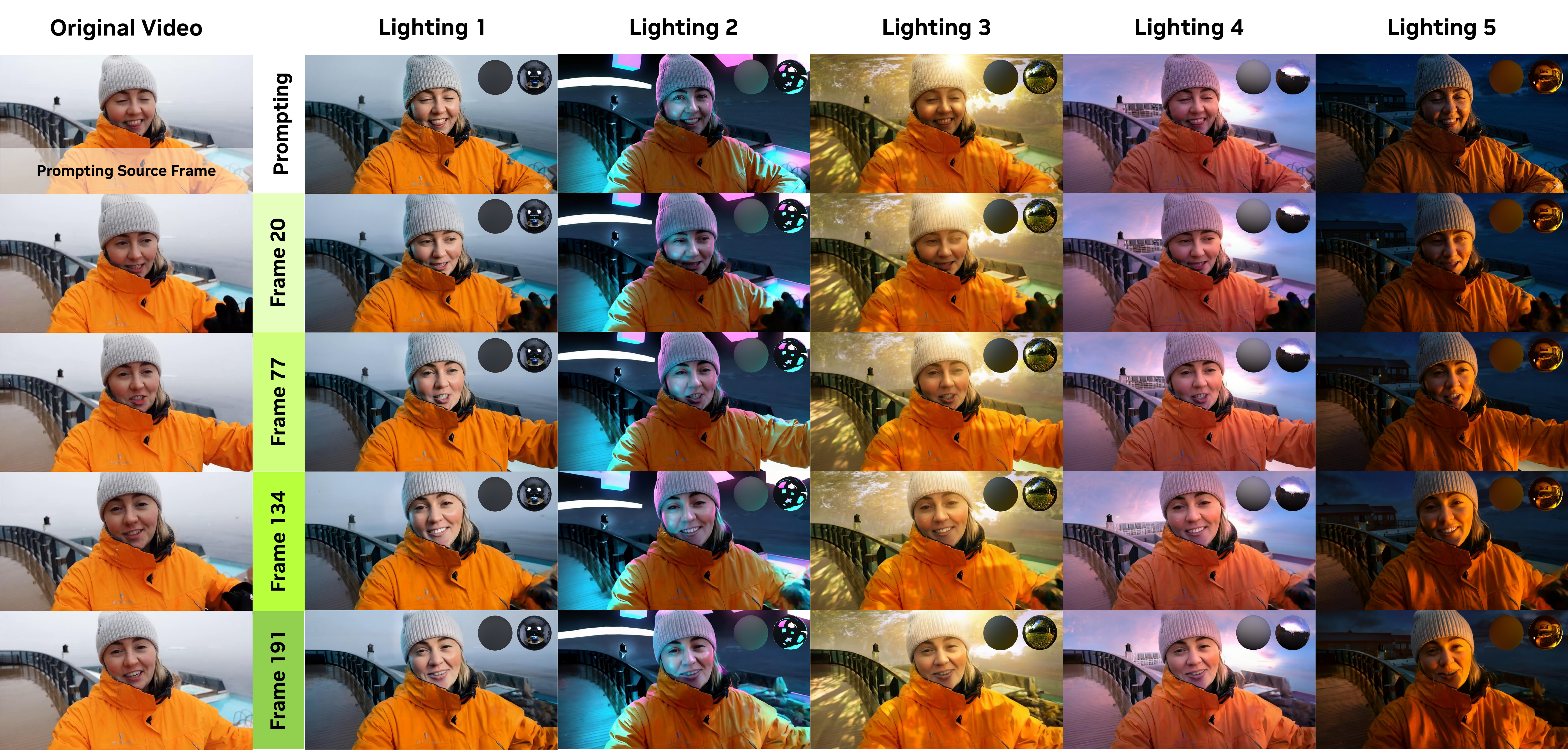}
    \caption{\textbf{Long-horizon relighting with warm-start prompting.} We use Nano Banana Pro to generate a starter image that initializes the relighting under the target illumination, after which the relit sequence is produced by chained inference across chunks. We visualize frames sampled at the same relative position from consecutive chunks to highlight temporal consistency under different lighting conditions.}
    \vspace{-10pt}
    \label{fig:qualitative_prompting_relighting}
\end{figure}

\section{Experiments}
\begin{figure}[t!]
    \centering
    \includegraphics[width=\linewidth,trim=0 825 0 0,clip]{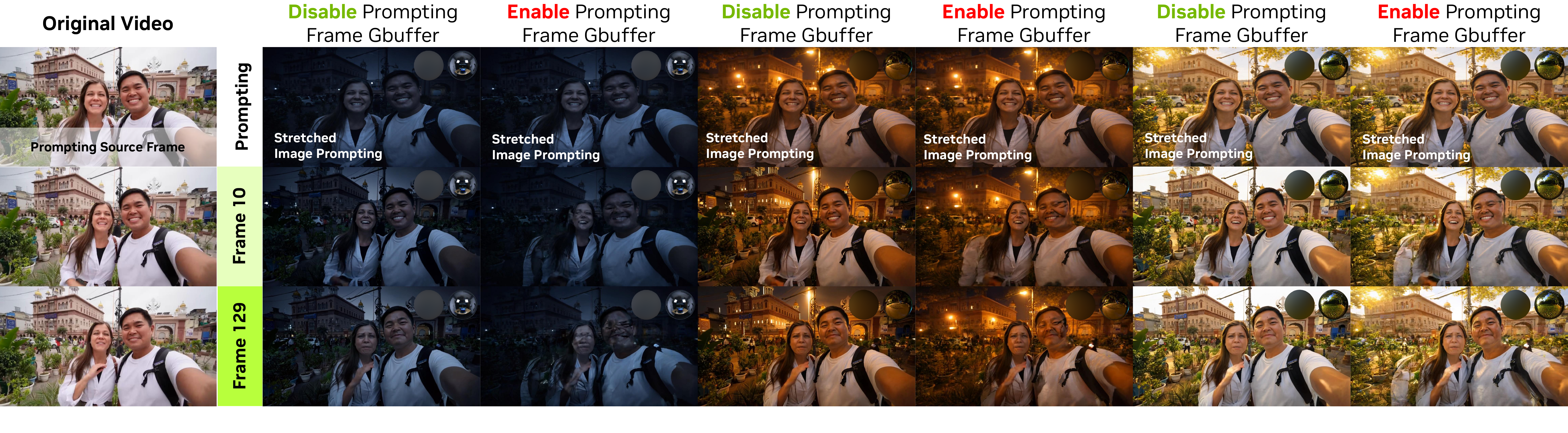}
    \caption{\textbf{Ablation on stretched warm-start prompting.} When the starter image is not well aligned with the predicted structural cues, directly conditioning on the corresponding G-buffer can introduce blur or content mixing. Disabling the warm-start G-buffer for the prompted frame preserves the initialization benefit while avoiding misaligned guidance.}
    \label{fig:qualitative_prompting_ablation}
\end{figure}

\begin{figure}[t]
    \centering
    \includegraphics[width=\linewidth,trim=0 650 400 0,clip]{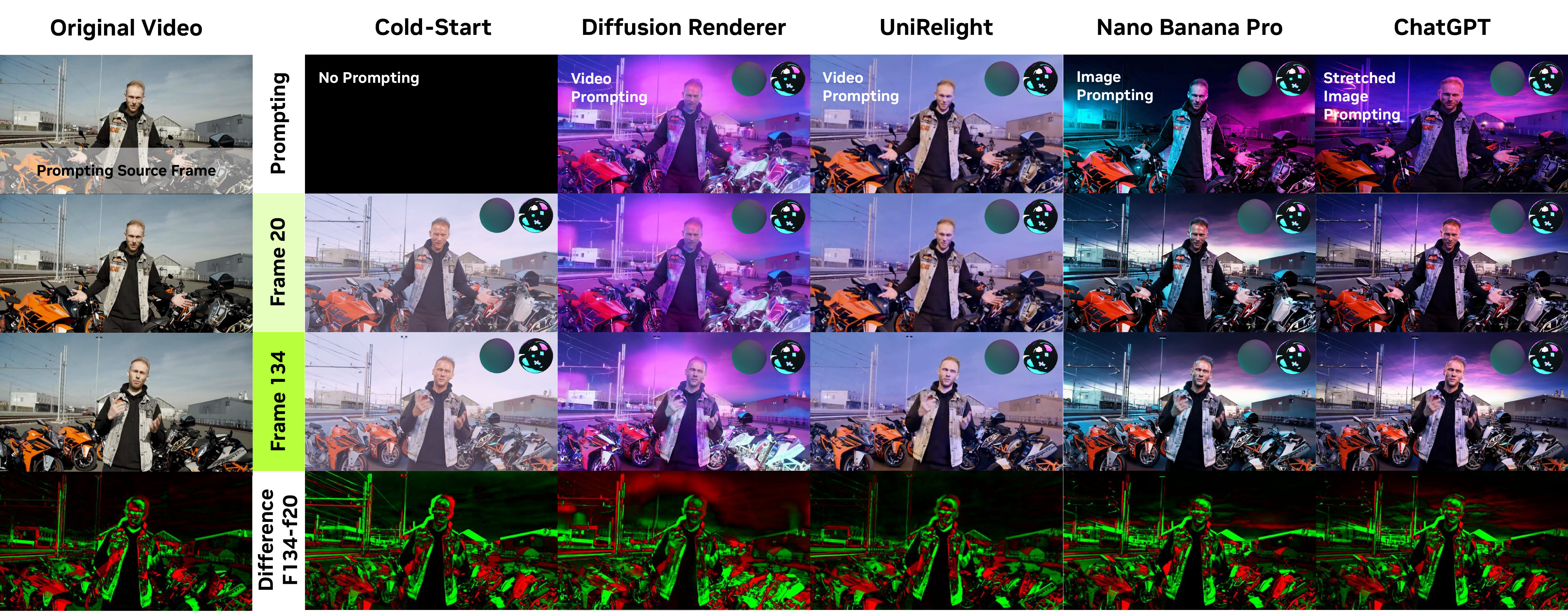}
    \caption{\textbf{Generality of warm-start prompting.} Our framework supports multiple initialization modes, including cold start, video-based prompting, and image-based prompting. We compare frames across chunks to show that different prompt sources support stable long-horizon generation. Difference maps highlight true scene dynamics while remaining largely black in static regions, indicating preserved scene and lighting consistency across chunks.}
    \vspace{-10pt}
    \label{fig:qualitative_prompting_generality}
\end{figure}

\begin{figure}[t]
    \centering
    \includegraphics[width=\linewidth,trim=0 50 0 0,clip]{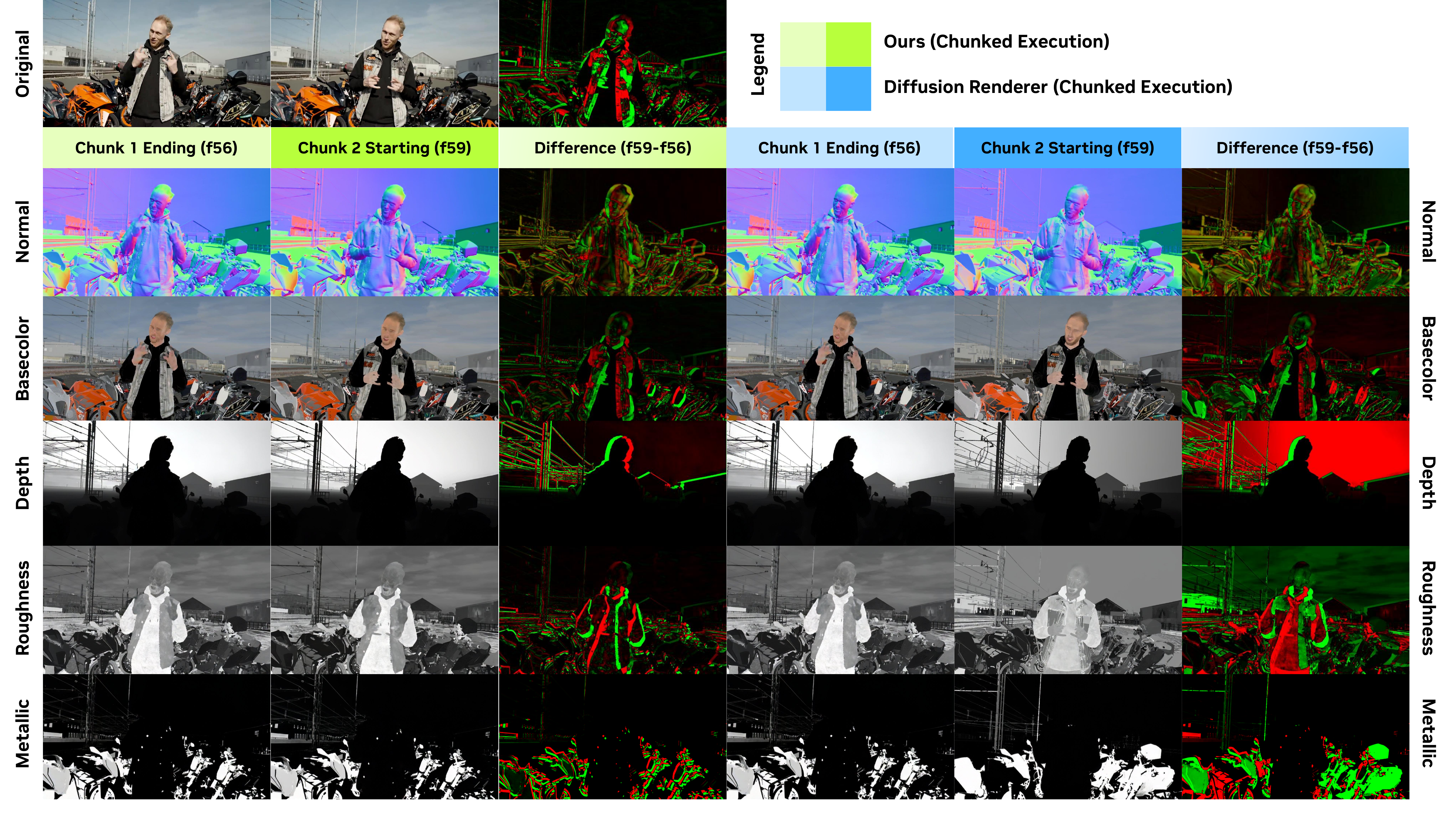}
    \caption{\textbf{Temporal consistency of long-horizon inverse decomposition.} We compare predictions around chunk boundaries using frame differences to reveal cross-chunk variation. Our method maintains clean, localized changes near true scene dynamics, indicating substantially improved inverse-stage consistency.}
    \vspace{-10pt}
    \label{fig:qualitative_temporal_consistency_inverse}
\end{figure}

\begin{figure}[t]
    \centering
    \includegraphics[width=\linewidth,trim=0 530 0 0,clip]{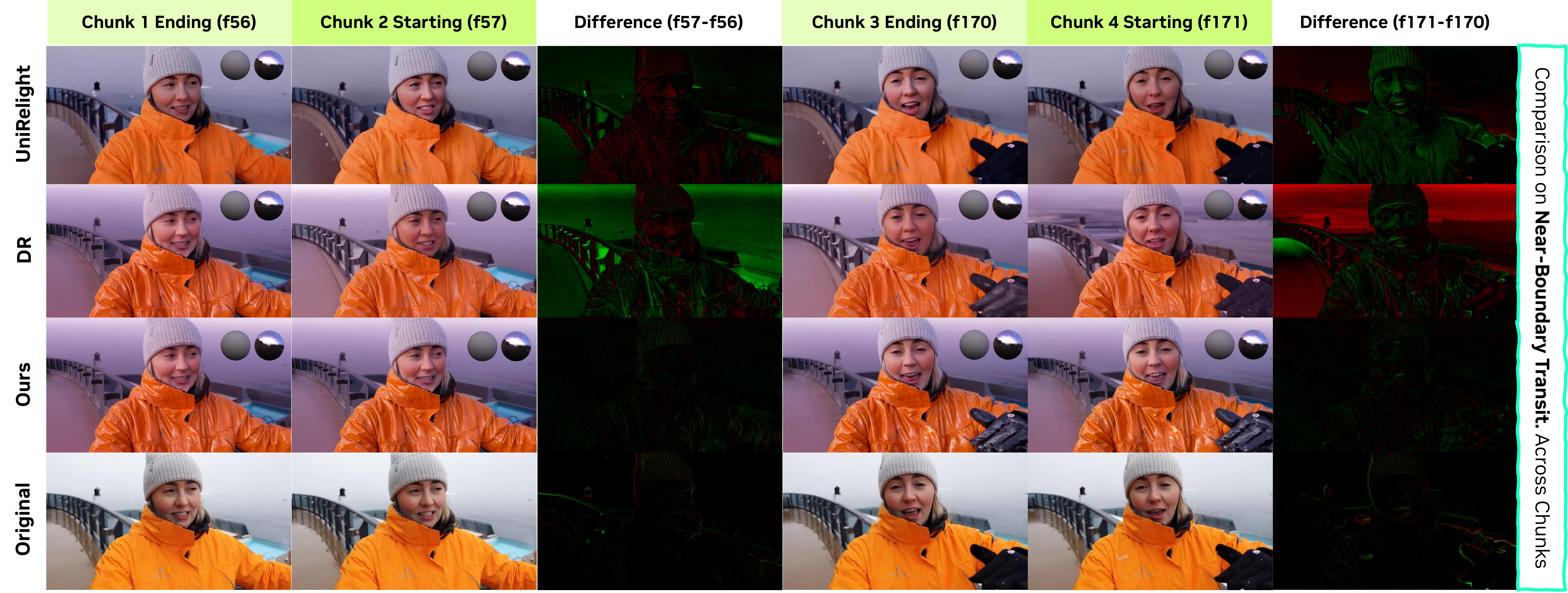}
    \includegraphics[width=\linewidth,trim=0 530 0 0,clip]{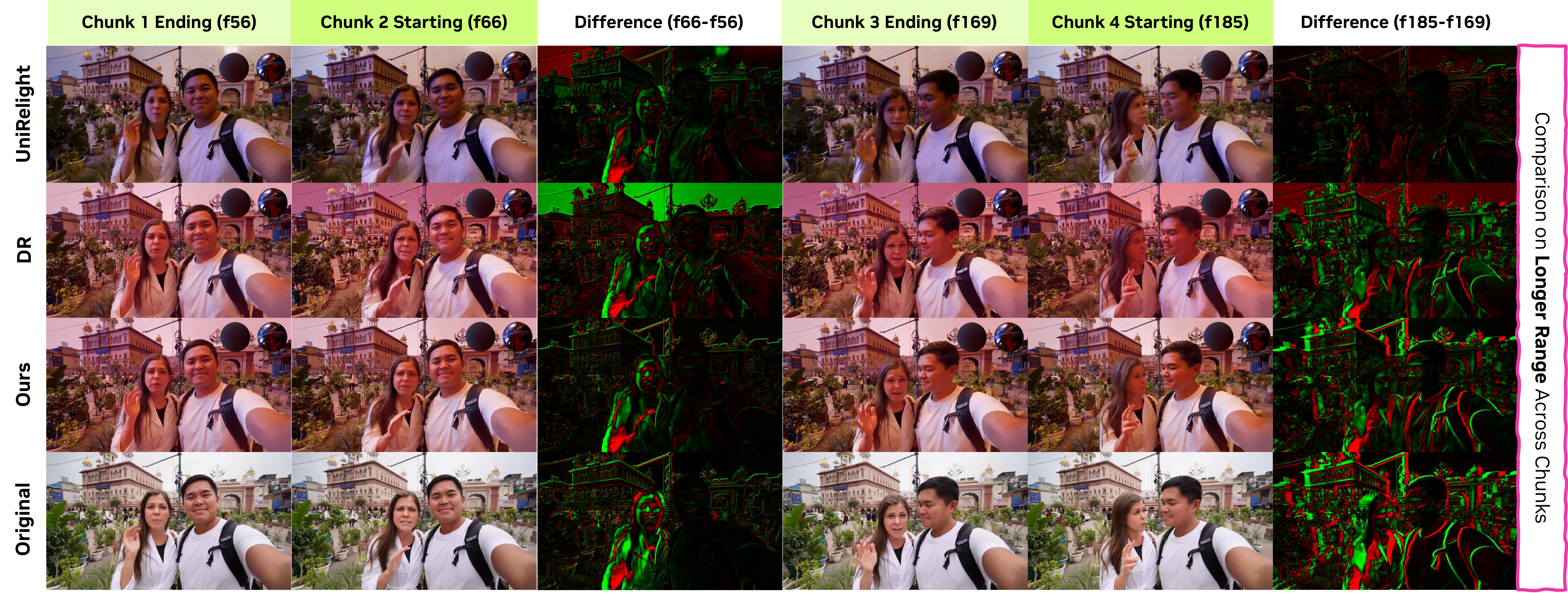}
    \caption{\textbf{Temporal consistency of long-horizon forward rendering.} We compare both near-boundary transitions and longer-range frame dynamics across chunks. Near-boundary comparisons should show only thin differences around true motion, while longer-range comparisons naturally produce thicker dynamic regions due to accumulated camera and object motion. Our results follow this expected pattern and remain close to the input-video difference maps, whereas the baselines exhibit broader changes in non-moving regions, indicating blur, flicker, and cross-chunk drift.}
    \vspace{-15pt}
    \label{fig:qualitative_temporal_consistency_forward}
\end{figure}

\begin{figure}[t]
    \centering
    \includegraphics[width=\linewidth,trim=0 200 300 0,clip]{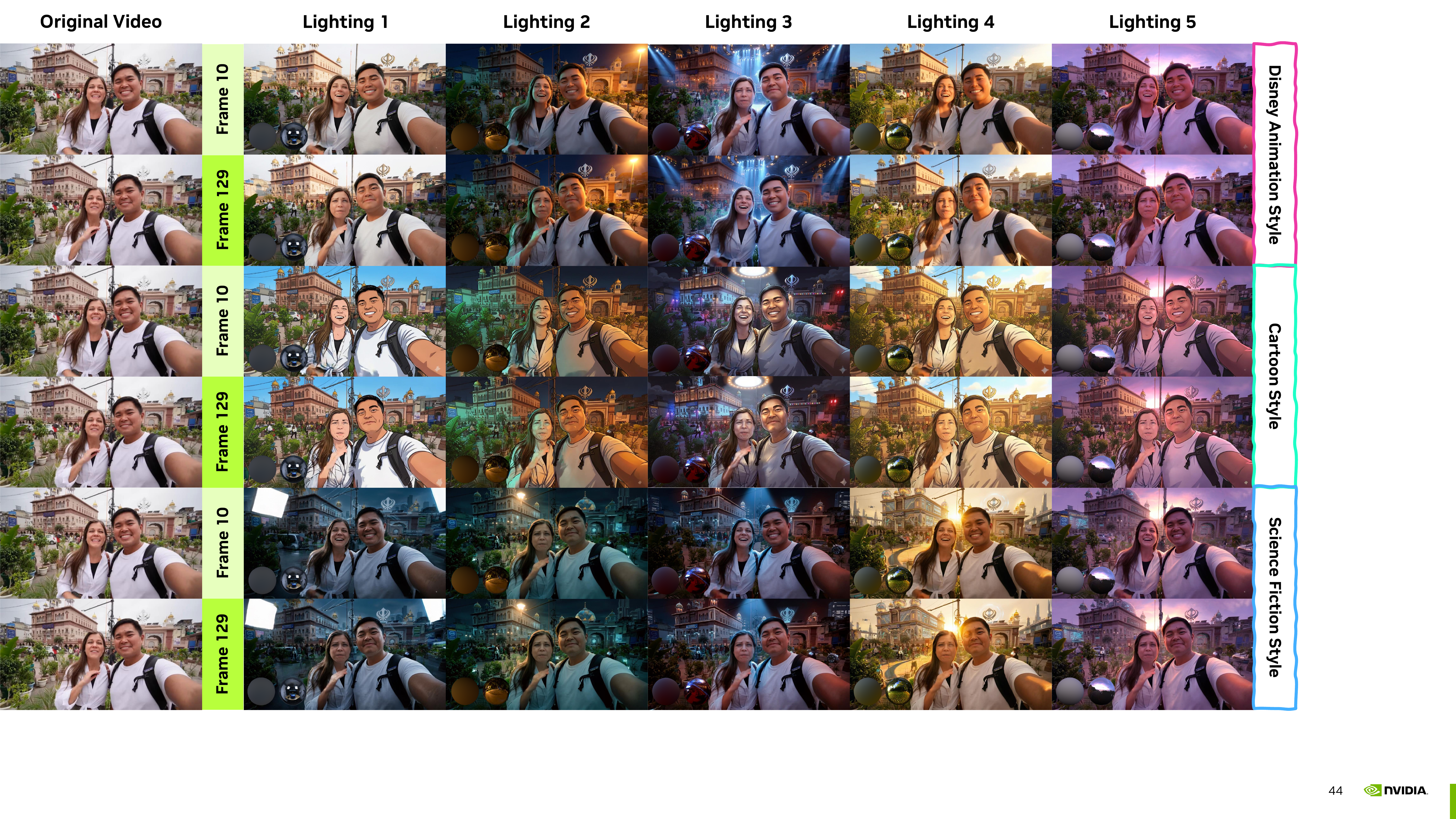}
    \vspace{-10pt}
    \caption{\textbf{Long-horizon editing.} The same warm-start mechanism can be used not only for relighting, but also for temporally consistent appearance and style edits over long-horizon videos.}
    \vspace{-20pt}
    \label{fig:qualitative_prompting_editing}
\end{figure}

\subsection{Dataset}

\paragraph{Synthetic Dataset.}
For training and controlled evaluation, we use the same synthetic data generation setup as DiffusionRenderer~\cite{DiffusionRenderer} and UniRelight~\cite{he2025unirelight}. The dataset is procedurally rendered from scenes built with 36.5K Objaverse LVIS objects, 4,260 PBR materials, and 766 HDRI environment maps, with randomized camera/object/lighting motion and additional primitives to increase geometric and lighting diversity. In our rendering pipeline, each video clip is rendered as a 121-frame sequence. For training, we follow the DiffusionRenderer/UniRelight convention and use the first 57 frames from each clip.

\paragraph{In-the-wild YouTube C.C.}
To evaluate generalization beyond synthetic data, we additionally curate an in-the-wild benchmark from YouTube Creative Commons videos. We collect around 100 arbitrary samples using broad keywords such as ``vlog,'' ``travel long take,'' and ``film long take.'' The resulting set covers diverse camera motion, object motion, and scene layouts, and is used primarily for qualitative evaluation of long-horizon temporal consistency and warm-start behavior under practical deployment conditions.

\subsection{Qualitative}

\paragraph{Relighting Long-horizon Videos.}
We present relighting results in Fig.~\ref{fig:qualitative_prompting_relighting}, where the warm-start prompting uses a high-quality image generated by Nano Banana Pro~\cite{google2026nanobanana} that is spatially consistent with the input frame. To obtain this warm-start image, we use the first frame of the input video and the target HDRI as inputs to Nano Banana Pro, with the prompt: ``\textit{Relight the first image using the second image as the environment lighting. Update the lighting for both the foreground and background, while preserving the content, resolution, and aspect ratio of the first image. Use the second image only for lighting, and do not transfer its content.}'' This warm-start image provides an explicit target-domain anchor for the first chunk, after which our model maintains the relit appearance through chained latent propagation across later chunks. As shown in Fig.~\ref{fig:qualitative_prompting_relighting}, this produces visually stable relighting across long horizons, with consistent shading and appearance preserved across consecutive chunks under the same target illumination.

In practice, however, the warm-start image often does not exactly match the input frame layout. The external editor may slightly change the framing, or return a stretched image that no longer matches the predicted G-buffer intrinsics for that frame. We therefore study this practical case in Fig.~\ref{fig:qualitative_prompting_ablation}. Even under this mismatch, the warm-start image can still provide a useful target-domain initialization. The main issue is the conflict between the prompted RGB image and the predicted G-buffer intrinsics used by the forward stage. If both are used together when they are spatially inconsistent, the model receives incompatible cues, which can lead to blur or content mixing. Disabling the warm-start G-buffer for that frame removes this conflict and allows the model to retain the benefit of the appearance prior. This shows that warm-start prompting remains effective even when the starter image is imperfect.

We then examine the generality of the warm-start choice in Fig.~\ref{fig:qualitative_prompting_generality}. Our framework is not tied to a single initialization source: besides the standard warm-start image, it also supports cold start, video-based prompting, and alternative image-based prompts. Cold start is feasible, but without an explicit target-domain anchor it relies more heavily on the relighting prior learned from the training data, which can weaken strong relighting cues such as specular detail. As shown by the ablation in Fig.~\ref{fig:qualitative_prompting_ablation}, misalignment between the prompt and the predicted G-buffer intrinsics can introduce conflicting cues. This makes video prompting less flexible in practice, since it requires stronger spatial alignment with the target sequence, whereas image-based prompting is more robust because any mismatch is confined to the warm-start frame and can be mitigated by disabling the corresponding G-buffer. In Fig.~\ref{fig:qualitative_prompting_generality}, we further show that warm-start images generated by both Nano Banana~\cite{google2026nanobanana} and ChatGPT 4o~\cite{openai2025gpt4o_imagegen} can be used within the same framework.

\paragraph{Cross-Chunk Consistency.}
We next evaluate the temporal consistency enabled by the chained design. Fig.~\ref{fig:qualitative_temporal_consistency_inverse} shows consistency in the inverse stage, and Fig.~\ref{fig:qualitative_temporal_consistency_forward} shows consistency in the forward relighting stage, including both near-boundary transitions and longer-range comparisons across chunks.

We compare against DiffusionRenderer (DR)~\cite{DiffusionRenderer} and UniRelight~\cite{he2025unirelight} on in-the-wild YouTube samples. Because chunk-boundary discontinuities are the dominant failure mode in long-horizon chunked inference, we visualize frame differences to determine whether each method captures true scene dynamics or introduces chunk-specific appearance changes. Ideally, the difference maps should respond mainly in regions of actual motion, while non-moving regions remain largely black. Responses outside these dynamic regions indicate temporal drift or inconsistent re-inference across chunks.

In the inverse stage (Fig.~\ref{fig:qualitative_temporal_consistency_inverse}), the baselines often show broader appearance changes across chunk boundaries, including responses in regions that should remain static. This indicates that scene factors are being re-estimated inconsistently from one chunk to the next. By contrast, our method keeps the differences concentrated near the true motion regions, while the rest of the frame remains much more stable. This indicates reduced cross-chunk drift in the inverse predictions. For the inverse stage, our method uses DiffusionRenderer to initialize the first chunk. The consistency gain observed at later chunk boundaries therefore comes from the chained propagation across subsequent chunks, rather than from a different initialization at the start.

The same pattern becomes even clearer in the forward stage (Fig.~\ref{fig:qualitative_temporal_consistency_forward}), where we compare both near-boundary transitions and longer-range dynamics across chunks. In the near-boundary comparisons, the motion between adjacent frames is usually very small, so the difference maps should show only thin boundaries around moving objects or slight camera motion, with the rest of the frame remaining black. Our results follow this expected pattern and closely match the input-video difference maps, while the baselines already show broader changes in non-moving regions. In the longer-range comparisons, the accumulated motion from camera movement and object motion becomes more visible, so the dynamic regions in the difference maps are naturally thicker. Again, our results remain consistent with the input-video difference maps by highlighting these true dynamic regions, while the baselines continue to respond strongly in areas that should remain stable. This leads to blur, flicker, and appearance drift. Overall, these results show that our method captures the true scene dynamics while maintaining scene and lighting consistency in non-moving regions across both short- and long-range comparisons.

\paragraph{Long-horizon Editing.}
Finally, Fig.~\ref{fig:qualitative_prompting_editing} shows that the same warm-start prompting mechanism supports a broader class of long-horizon generation tasks beyond relighting. When the warm-start reference is aligned with the input structural conditions, the prompted appearance can be preserved and propagated throughout the sequence. This remains effective even when the prompted appearance is not explicitly represented in the training data. Rather than reproducing only familiar object identities or styles, the model learns to extend the prompted target-domain distribution over time while maintaining temporal consistency in scene content and lighting. As a result, the same mechanism can animate a single edited image into a coherent long-horizon sequence, enabling temporally consistent style transfer, content editing, and relighting within one unified framework.

This behavior is especially relevant to practical film production. In many studio workflows, scenes are captured under controlled conditions such as green-screen or blue-screen setups, while much of the final look is determined later during visual effects and lighting design. In this setting, our warm-start prompting provides a natural way to inject a desired appearance at the beginning of the sequence and propagate it consistently across time. Figure~\ref{fig:qualitative_prompting_editing} illustrates this capability with examples of both subject-level edits and global style changes that remain coherent throughout the long-horizon videos.

\begin{figure}[t]
    \centering
    \includegraphics[width=\linewidth,trim=0 1000 0 0,clip]{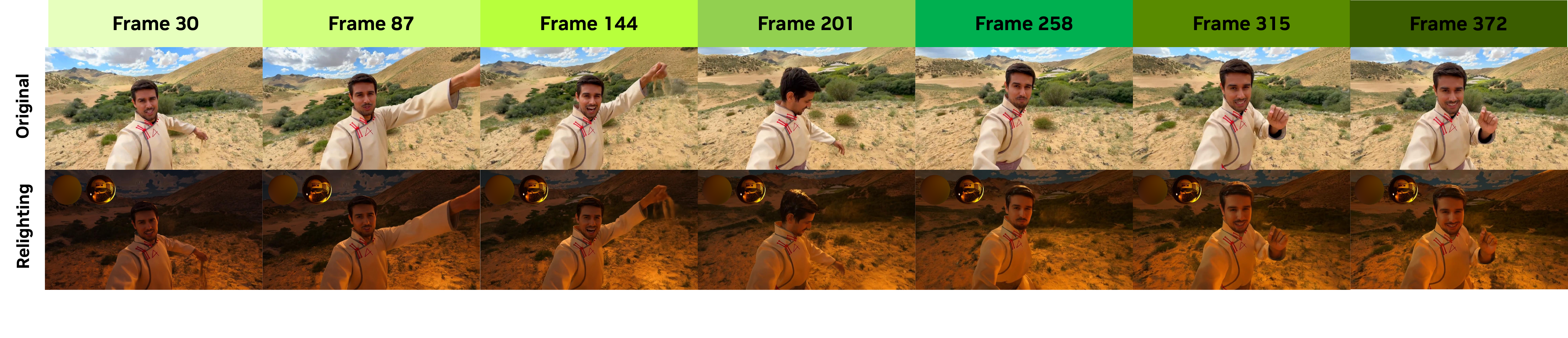}
    \vspace{-20pt}
    \caption{\textbf{Long-horizon limitations.} Beyond sufficiently long chains, the relighting remains preserved, but scene content gradually loses fine details.}
    \vspace{-15pt}
    \label{fig:qualitative_limitation}
\end{figure}

\paragraph{Limitations.}
As visualized in Fig.~\ref{fig:qualitative_limitation}, a limitation is fine-detail degradation at extremely long horizons. While chained propagation improves relighting consistency over long-horizon videos, the output quality remains bounded by the learned appearance prior. In our experiments, beyond roughly five chunks (about 300 frames), relighting stays stable, but scene content gradually loses high-frequency details. Extending the usable horizon will likely require training data and supervision targets that better cover the long-horizon appearance space encountered in practice.

\subsection{Quantitative}

We quantitatively evaluate cross-chunk temporal consistency for both inverse decomposition and forward relighting on the synthetic dataset. For this evaluation, we use the last 57 frames of each rendered 121-frame video in the testing set. To construct a controlled cross-chunk setting, we form two consecutive 57-frame chunks by taking the selected 57-frame sequence and concatenating it with its duplicated reversed copy. Under this construction, the overlapping frames across the chunk boundary are exactly identical in content, so any prediction difference across the boundary directly reflects temporal inconsistency introduced by chunked inference rather than true scene motion.

We report mean squared error (MSE) in two ways to complement the qualitative analysis. \emph{Boundary} MSE is computed on the overlapping frames across a chunk transition, directly measuring boundary consistency. \emph{Sequence} MSE is computed over the full predicted chunk, measuring how inconsistency propagates beyond the overlap. We evaluate both inverse decomposition outputs and the final relighting output. Table~\ref{tab:quantitative} shows that our method consistently improves both metrics across all inverse G-buffer prediction and the final relighting output.

\begin{table}[t]
\centering
\caption{\textbf{Quantitative temporal consistency on the synthetic dataset.} We report MSE under the controlled repeated-chunk setting. \emph{Boundary} is computed only on the overlapping frames across the chunk boundary, while \emph{Sequence} is computed over the full predicted chunk. Lower is better.}
\label{tab:quantitative}
\setlength{\tabcolsep}{3.5pt}
\renewcommand{\arraystretch}{1.05}
\begingroup
\fontsize{7}{8}\selectfont
\begin{tabular}{llccccccc}
\toprule
Method & Scope & Normal & BaseColor & Depth & Roughness & Metallic & Specular & Relighting \\
\midrule
DR~\cite{DiffusionRenderer} & Boundary & 0.1379 & 0.1613 & 0.1576 & 0.2614 & 0.2050 & 0.1205 & 0.1048 \\
Ours                        & Boundary & \textbf{0.0971} & \textbf{0.1173} & \textbf{0.1181} & \textbf{0.1042} & \textbf{0.0542} & \textbf{0.0935} & \textbf{0.0729} \\
\midrule
DR~\cite{DiffusionRenderer} & Sequence & 0.1425 & 0.1670 & 0.2208 & 0.2664 & 0.2409 & 0.1242 & 0.1063 \\
Ours                        & Sequence & \textbf{0.1007} & \textbf{0.1325} & \textbf{0.1383} & \textbf{0.1199} & \textbf{0.1088} & \textbf{0.1137} & \textbf{0.0906} \\
\bottomrule
\end{tabular}
\endgroup
\vspace{-10pt}
\end{table}

\subsection{Implementation Details}
Our models are initialized from \textit{Cosmos-Predict1-7B-Video2World}~\cite{agarwal2025cosmos} pretrained weights and fine-tuned on our relighting training data using a Cosmos-based DiT video diffusion backbone. We train the forward rendering (relighting) model and the inverse decomposition model separately under the same hardware setup: the forward model is trained for approximately one month on 16 NVIDIA H100 GPUs, and the inverse model is trained for approximately two weeks on 16 NVIDIA H100 GPUs. Training is performed in mixed precision (BF16) with AdamW and latent-space conditioning, where all inputs (video, intrinsics, and lighting cues) are encoded into tokens and concatenated with type embeddings.
We employ \emph{masked target-domain self-conditioning}: during training, we randomly sample temporal masks over the target-domain latents to expose the model to arbitrary partial-context continuation, while at inference we keep only the first frame unmasked and predict the remaining frames conditioned on this initial anchor.

\section{Conclusion}
We presented a long-horizon video relighting framework that improves cross-chunk consistency by reframing chunked execution as temporally conditioned latent domain translation, combining chained propagation with warm-start prompting. Experiments on both synthetic and in-the-wild videos show temporally consistent inverse decomposition for forward relighting. The main limitation is long-horizon detail degradation: while relighting remains stable, fine details weaken as the chain grows longer. Future work includes reducing this quality drop and exploring long-horizon free-form creative editing via prompt anchors.


\clearpage
\begingroup
\thispagestyle{empty}
\AddToShipoutPictureBG*{%
  \AtPageCenter{%
    \makebox(0,0)[c]{\includegraphics[page=1,width=\paperwidth,height=\paperheight,keepaspectratio]{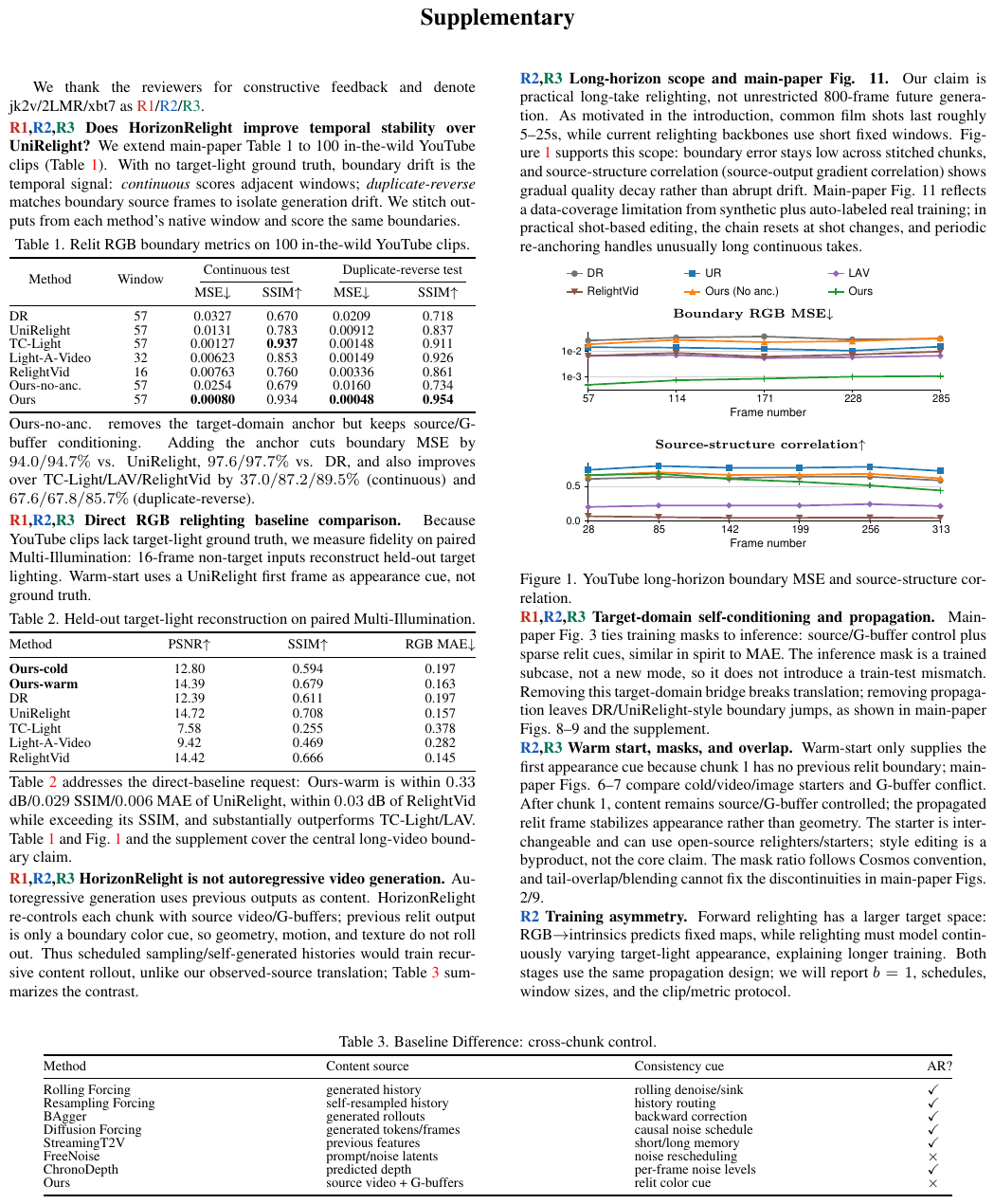}}%
  }%
}
\null
\clearpage
\endgroup

\clearpage
%
%
\bibliographystyle{splncs04}
\bibliography{main}
\end{document}